\documentclass[conference]{IEEEtran}
\IEEEoverridecommandlockouts

\usepackage{cite}
\usepackage{amsmath,amssymb,amsfonts}
\usepackage{algorithmic}
\usepackage{graphicx}
\usepackage{textcomp}
\def\BibTeX{{\rm B\kern-.05em{\sc i\kern-.025em b}\kern-.08em
    T\kern-.1667em\lower.7ex\hbox{E}\kern-.125emX}}

\usepackage{booktabs}
\usepackage{amssymb}
\usepackage{amsthm}
\usepackage{mathtools}
\usepackage{microtype}
\usepackage{graphicx}
\usepackage{caption}
\usepackage{wrapfig}
\usepackage{booktabs}
\usepackage{hyperref}
\usepackage{cleveref}
\usepackage[table]{xcolor}
\usepackage{multicol}
\usepackage[export]{adjustbox}
\usepackage{siunitx}
\usepackage{physics}
\usepackage{subcaption}
\usepackage{balance}
\usepackage{makecell}       % table linebreak
\usepackage{pifont}         % checkmarks
\usepackage{array}          % for column alignment customization
\usepackage[acronym]{glossaries}

\newsavebox{\twosubbox}

\newcommand{\x}{{\mathbf x}}
\newcommand{\y}{{\mathbf y}}

\newcommand{\w}{{\mathbf w}}
\newcommand{\n}{{\mathbf n}}

\newcommand{\A}{{\mathbf A}}
\newcommand{\I}{{\mathbf I}}

\newcommand{\redcross}{{\textcolor{red}{\ding{55}}}}
\newcommand{\greencheck}{{\textcolor{green}{\ding{51}}}}
\newcommand{\orangetilde}{{\textcolor{orange}{$\mathbf{\sim}$}}}

\definecolor{green}{RGB}{76, 153, 0}
\definecolor{red}{RGB}{204, 0, 0}
\definecolor{orange}{RGB}{255, 165, 0}
\definecolor{purple}{RGB}{180, 50, 177}

\newacronym{awgn}{AWGN}{additive white Gaussian noise}
\newacronym{dgms}{DGMs}{Deep Generative Models}
\newacronym{dgm}{DGM}{Deep Generative Model}
\newacronym{dps}{DPS}{Diffusion Posterior Sampling}
\newacronym{dms}{DMs}{Diffusion Models}

\begin{document}

\title{Sequential Posterior Sampling with Diffusion Models}

\author{\begin{tabular}{c}Tristan S.W. Stevens$^{\star}$, Oisín Nolan$^{\star}$, Jean-Luc Robert$^{\dagger}$, Ruud J.G. van Sloun$^{\star}$\end{tabular} \vspace{0.2cm}\\
\IEEEauthorblockA{
    $^{\star}$Dept. of Electrical Engineering, Eindhoven University of Technology, The Netherlands \\
    $^{\dagger}$Philips Research North America, Cambridge MA, USA}
}

\maketitle

\begin{abstract}
Diffusion models have quickly risen in popularity for their ability to model complex distributions and perform effective posterior sampling. Unfortunately, the iterative nature of these generative models makes them computationally expensive and unsuitable for real-time sequential inverse problems such as ultrasound imaging. Considering the strong temporal structure across sequences of frames, we propose a novel approach that models the transition dynamics to improve the efficiency of sequential diffusion posterior sampling in conditional image synthesis. Through modeling sequence data using a video vision transformer (ViViT) transition model based on previous diffusion outputs, we can initialize the reverse diffusion trajectory at a lower noise scale, greatly reducing the number of iterations required for convergence. We demonstrate the effectiveness of our approach on a real-world dataset of high frame rate cardiac ultrasound images and show that it achieves the same performance as a full diffusion trajectory while accelerating inference 25$\times$, enabling real-time posterior sampling. Furthermore, we show that the addition of a transition model improves the PSNR up to 8\% in cases with severe motion. Our method opens up new possibilities for real-time applications of diffusion models in imaging and other domains requiring real-time inference.
\end{abstract}
\section{Introduction}
\label{sec:intro}

Deep generative models are celebrated for their ability to model complex distributions. Their use in inverse problem solving has unlocked new applications involving high-dimensional data. \acrfull{dms} are particularly attractive generative models due to their interpretable denoising score matching objective and stable sampling procedure. Despite these benefits, the iterative nature of sampling from prior and posterior distributions with diffusion models inhibits their use in demanding real-time imaging applications with high data-rates such as cardiac ultrasound \cite{stojanovski2023echo, stevens2024dehazing} or automotive radar \cite{wu2024diffradar, overdevest2024model}.

There have been several works on accelerating \acrshort{dms}. These can be roughly categorized in two lines of research. On the training end,  \cite{salimans2021progressive} proposes a \emph{progressive distillation} method that augments the training of the \acrshort{dms} with a student-teacher model setup. In doing this, they are able to drastically reduce the number of sampling steps. Some methods aim to execute the the diffusion process in a reduced space to accelerate the diffusion process. While \cite{jing2022subspace} restricts diffusion through projections onto subspaces, \cite{vahdat2021score} and \cite{rombach2022high} run the diffusion in the latent space. On the other side of the spectrum, the sampling procedure itself can be altered. Inspired by momentum methods in sampling, \cite{daras2022soft} introduces a momentum sampler for \acrshort{dms}, which leads to increased sample quality with fewer function evaluations. More related to this work is a sampling strategy known as \emph{Come-Closer-Diffuse-Faster} (CCDF)~\cite{chung2022come}, which leverages a neural network based estimate of the posterior mean to reduce the number of reverse diffusion steps needed. Nonetheless, CCDF and the other aforementioned methods do not exploit the temporal structure across frames in sequential data which we demonstrate improves the solvability of inverse problems.

Video diffusion models extent on previous works by training a diffusion prior jointly on a sequence of frames \cite{ho2022video}.  While they have been extensively explored for tasks such as  \emph{text-to-video}~\cite{ho2022imagen} and \emph{image-to-video}~\cite{ni2023conditional} generation, there has been limited research on their application to video reconstruction tasks. Some works have investigated the use of \acrshort{dms} for time-series; \cite{rasul2021autoregressive}, for example, proposes a conditional diffusion model for time series forecasting. However, these works do not consider the temporal structure across frames for accelerating the sampling process, rendering them too slow for real-time inference.

In this work, we propose a novel autoregressive method for initializing successive diffusion trajectories for reconstruction of sequence data. We provide two flavors named \emph{SeqDiff} and \emph{SeqDiff+} which both leverage the temporal correlation across frames, by using the diffusion model output of previous frames as a starting point for the current posterior sampling procedure. SeqDiff straightforwardly initializes with the previous frame, which we show is often reasonable given high frame rates. Expanding on this idea, SeqDiff+ specifically models the transition between subsequent frames using a \emph{Video Vision Transformer} (ViViT)~\cite{arnab2021vivit} for a more accurate initialization, mitigating the effect of severe motion across frames.

To evaluate our method, we turn to echocardiography, which is the imaging of the heart using medical ultrasound. The real-time nature and high data-rates resulting from this sensory data encapsulate the challenges targeted by the proposed method. \acrshort{dms} have been effectively applied to cardiac ultrasound, from removing multipath scattering (dehazing) \cite{stevens2024dehazing} to segmentation~\cite{stojanovski2023echo} and beyond. However, accurate and fast image reconstruction using \acrshort{dms} remains a challenge.

Our main contributions can be summarized as follows:
\begin{itemize}
    \item We propose autoregressive tracking of posterior samples across the noise manifolds in diffusion models to accelerate reconstruction of sequential data.
    \item We provide two variants, SeqDiff and SeqDiff+, both of which rely on previous diffusion posterior estimates for initialization. SeqDiff+ further leverages a Video Vision Transformer to model the transitions between frames.
    \item We evaluate our method on compressed sensing echocardiography, showing that our method improves image quality while accelerating the sampling process.
\end{itemize}
The remainder of this paper is organized as follows. In Section~\ref{sec:background} we provide background on both posterior sampling with \acrshort{dms} as well as sequence modeling. In Section~\ref{sec:methods} we proceed with introduction of our methods, which are subsequently evaluated and concluded in sections~\ref{sec:results} and \ref{sec:conclusions}, respectively.

\section{Background}
\label{sec:background}
\subsection{Diffusion Models}
\acrfull{dms} are a class of probabilistic generative models that learn the reversal of a forward corruption process, which add progressively increasing levels of Gaussian noise until the data $\x_0 \equiv \x \sim p(\x)$ is transformed into a base distribution $\x_{\mathcal{T}} \sim \mathcal{N}(0, \I)$. The continuous forward process $\x_0\rightarrow\x_\tau \rightarrow\x_\mathcal{T}$, with diffusion time $\tau\in\left[0, \mathcal{T}\right]$ can be formally described by a variance preserving stochastic differential equation (VP-SDE) \cite{song2020score} $\mathrm{d} \x = -\frac{1}{2}\beta(\tau)\x\mathrm{d}\tau + \sqrt{\beta(\tau)}\mathrm{d}\w$, where $\beta(\tau)$ is the noise schedule, and $\w$ a standard Wiener process. Diffused samples from $p(\x_\tau\vert\x_0)=\mathcal{N}(\alpha_\tau \x_0, \sigma_\tau^2 \I)$ can be directly generated by the following parameterization:
\begin{equation}
    \x_\tau = \alpha_\tau \x_0 + \sigma_\tau \mathbf{\epsilon}, \quad \mathbf{\epsilon}\in\mathcal{N}(0, \I),
\end{equation}
where $\sigma_\tau = 1 - e^{-\int_0^\tau\beta(s)\mathrm{d}s}$ and $\alpha_\tau = \sqrt{1 - \sigma_\tau^2}$ are the noise and signal rates, respectively. The objective of generative models is to generate samples from the distribution of interest given samples from some tractable distribution. Accordingly, a corresponding reverse-time SDE can be constructed to achieve this:
\begin{equation}
    \mathrm{d} \x = \left[-\frac{1}{2}\beta(\tau)\x - \beta(\tau) \nabla_{\x_\tau}\log{p(\x_\tau)}
    \right] \mathrm{d}\tau + \sqrt{\beta(\tau)}\mathrm{d}\bar{\w},
\label{eq:rev-sde}
\end{equation}
where $\mathrm{d} \tau$ and $\mathrm{d} \bar{\w}$ are now processes running backwards in diffusion time. From this reverse SDE the gradient of the log-likelihood of the data arises $\nabla_\x \log p_t(\x)$, also known as the \emph{score~function} which provides information on how to adjust $\x_\tau$ to move it towards $\x_0$ and can be modeled using neural network parameters $\theta$ leading to the following approximation: $s_\theta(\x_\tau, \tau) \approx \nabla_{\x_\tau}\log{p(\x_\tau)}$. As shown in \cite{vincent2011connection}, the score model $s_\theta$ can be learned with the denoising score matching objective
\begin{equation}
    \mathcal{L}(\theta) = \mathbb{E}_{\x_0\sim p(\x), \tau\sim\mathcal{U}[0, \mathcal{T}]}
    \left[
        \norm{
            s_\theta(\x_\tau, \tau) - \nabla_{\x_\tau}\log{p(\x_\tau\vert\x_0)}
        }_2^2
    \right],
\label{eq:dsm}
\end{equation}
which essentially trains a conditional denoising network at each diffusion timestep $\tau$. Finally, discretization of continuous process \eqref{eq:rev-sde} into $N$ equispaced diffusion steps is required to numerically approximate the reverse diffusion process and sample from the target distribution.

\begin{figure*}
    \centering
    \includegraphics[width=1\linewidth]{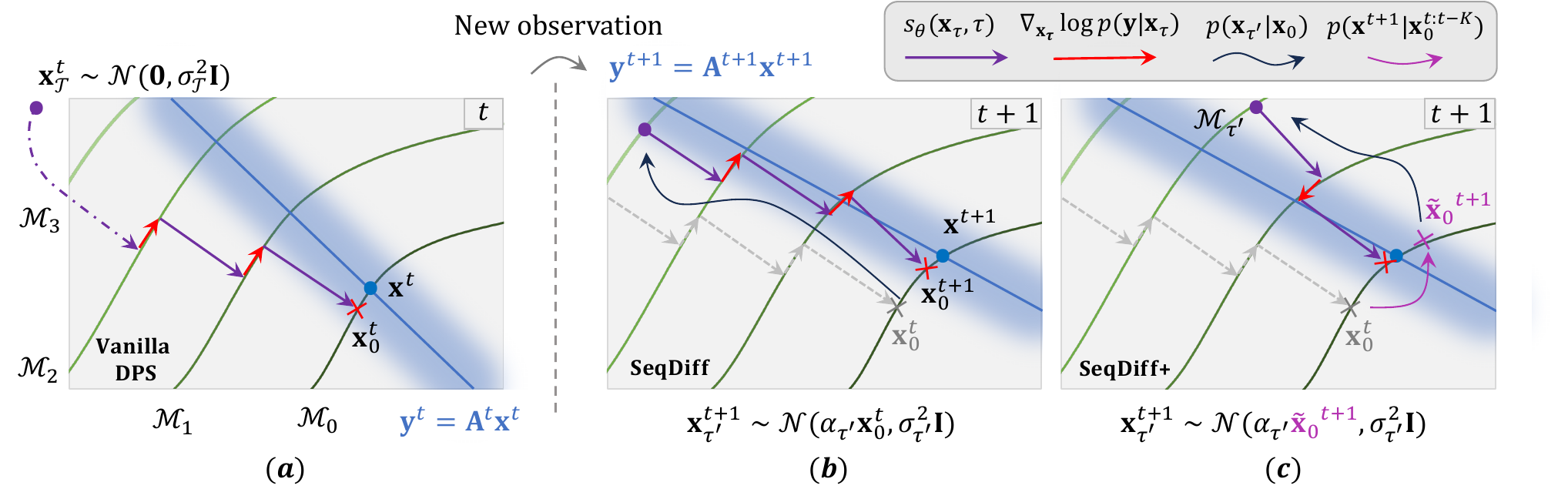}
    \captionsetup{skip=-4pt}
    \caption{Geometric representation of the reverse diffusion process and corresponding manifolds $\mathcal{M_\tau}$ for each diffusion timestep~$\tau$. In \textbf{(a)} a standard conditional reverse diffusion trajectory starting from a Gaussian sample $\x_\mathcal{T}\sim\mathcal{N}$ is shown with DPS as guidance rule \cite{chung2022diffusion}. For initialization of the next frame $t+1$, we propose two different methods SeqDiff and SeqDiff+, depicted in \textbf{(b)} and \textbf{(c)} respectively. In the first option we initialize the trajectory from a noised version of the Tweedie estimate of the previous frame, $p(\x_{\tau^\prime}^{t+1}|\x_0^{t})$ with $\tau^\prime \ll \mathcal{T}$. The second option improves upon this by predicting the next frame with $\tilde{\x}_0^{t+1} \approx f(\cdot)$, accounting for any motion between frames. This leads to the initialization $p(\x_{\tau^\prime}^{t+1}|\tilde{\x}_0^{t+1})$, with $\tau^\prime_{\text{SeqDiff+}} < \tau^\prime_{\text{SeqDiff}}$.}
    \label{fig:schematic}
    \vspace{-0.4cm}
\end{figure*}

\subsection{Posterior Sampling}
\label{sec:posterior}
Shifting our focus to inverse problems solving, which seeks to retrieve underlying signals $\x$ from corrupted observations $\y$, we can define a general linear forward model as follows:
\begin{equation}
    \y = \A \x + \n, \quad \y, \n \in \mathbb{R}^m, \x\in\mathbb{R}^n, \A \in \mathbb{R}^{m\times n}.
\label{eq:inverse}
\end{equation}
\acrshort{dms} can be extended to perform posterior sampling $p(\x_0\vert\y)$, through substitution of a conditional score into \eqref{eq:rev-sde}, which can be factorized into the pretrained score model and a noise perturbed likelihood score through Bayes' rule: $\nabla_{\x_\tau} \log p(\x_0\vert\y)\approx s_\theta(\x_\tau, \tau) + \nabla_{\x_\tau}\log{p(\y\vert\x_\tau})$. The intractability of the latter term has lead to several approaches to approximate it~\cite{song2023pseudoinverse, mardani2023variational}. Among the methods is Diffusion Posterior Sampling (DPS) \cite{chung2022diffusion}, which approximates the troubling $p(\x_0\vert\x_\tau)$, which leads to tractability of $p(\y\vert\x_\tau)$, as follows:
\begin{equation}
    p(\x_0\vert\x_\tau) \approx \mathbb{E}[\x_0\vert\x_\tau]\approx\frac{1}{\alpha_\tau}(\x_\tau + \sigma_\tau^2 s_\theta(\x_\tau, \tau))
\end{equation}
where the first approximation is substitution of the posterior mean for $\x_0$, and the second approximation the learned score model for the actual unconditional score function.

\subsection{Sequential inverse problems}
\label{sec:sequential}
In this work, we seek to address sequential inverse problems, also known as \emph{dynamic inverse problems}~\cite{hauptmann2021image}, which involve reconstructing from a sequence of time-dependent measurements $\y^t=\A^t\x^t + \n^t$ with a clear dependency between $\x^t$ and $\x^{t-1}$. To capture the intricate dynamics of temporal data, we look to sequence modeling which has become a fundamental task in applications such as speech recognition, natural language processing, and video analysis. We are interested in predicting future frames given past observations:
\begin{equation}
    p(\x^{t+1} \mid \x^{t}, \x^{t-1}, ...,  \x^{t-K}),
\end{equation}
where $K$ is the context window size. In the context of cardiac ultrasound this would translate to predicting a future frame given $K$ past frames. Traditional approaches to modeling sequences include hidden Markov models (HMMs), recurrent neural networks (RNNs), amongst which convolutional LSTMs (ConvLSTMs) \cite{shi2015convolutional} which have proven to work well for spatio-temporal data. More recently, transformer models have excelled especially in natural language processing tasks through self-attention mechanisms that capture long-range dependencies. The Video Vision Transformer (ViViT)~\cite{arnab2021vivit} extends this capability to video data by treating a stack of subsequent frames. Specifically, ViViTs extract non-overlapping, spatio-temporal tubes (3D patches), also known as tubelet embeddings, to tokenize the input video and accordingly process using multi-headed self-attention blocks.

\section{Methods}
\label{sec:methods}
The temporal correlation across subsequent frames can be heavily exploited to accelerate sequential posterior sampling $p(\x^t | \y^t, \x^{t-K:t-1})$ using \acrshort{dms}. We propose two techniques to initialize the reverse diffusion process corresponding to the current frame based on past observations in an efficient manner. In other words, given the diffusion posterior samples $\x_0$ of past frames $\left\{\x_0^{t}, \x_0^{t-1}, ...,  \x_0^{t-K}\right\}$ we would like to estimate $p(\x^{t+1}\mid \x_0^{t-K:t})$ such that the number of diffusion steps necessary is minimized. Since this is again a complex distribution, we instead estimate $p(\x^{t+1}_{\tau^\prime} \mid \x_0^{t-K:t})$, and assume it follows a tractable Gaussian with diagonal covariance. The challenge is to estimate the parameters of this distribution, as well as the diffusion time point $\tau^\prime$ for which the Gaussian approximation is accurate. We define the initialization diffusion scale as $\tau^\prime$ which lies somewhere on the diffusion timeline $0<\tau^\prime \ll \mathcal{T}$. Rather than starting each diffusion trajectory from scratch at $\tau=\mathcal{T}$ with a Gaussian sample $\x_\mathcal{T}\sim\mathcal{N}(\mathbf{0}, \sigma^2_\mathcal{T}\I)$, we use an appropriate estimate $\tilde{\x}$ based on past observations which we can diffuse forward up to $\tau=\tau^\prime$. The initialization of the (shortened) diffusion trajectory then becomes $\x_{\tau^\prime}\sim \mathcal{N}(\alpha_{\tau^\prime}\tilde{\x}, \sigma^2_{\tau^\prime}\I)$. For the discretized case, this reduces the number of steps to $N^\prime \ll N$, with $N^\prime=N\tau^\prime / \mathcal{T}$.

\begin{table}
    \footnotesize
    \renewcommand{\arraystretch}{1.1} % Increase row height for this table
    \newcolumntype{C}[1]{>{\centering\arraybackslash}p{#1}}
    \rowcolors{2}{gray!15}{white}
    \centering
    \begin{tabular}{l|C{1.7cm} C{1.7cm}| c}
        \toprule
            & \multicolumn{2}{c|}{\makecell{\textbf{Initialization} $p(\x_{\tau^\prime}^{t+1} \mid \x_0^{t+1})$\\$\mathcal{N(\quad\quad\mu\quad\quad, \quad\quad\sigma\quad\quad)}$}} & \makecell{\textbf{Sequence} \\ \textbf{modeling}} \\
        \midrule
        \textbf{Vanilla DPS}     & $\mathbf{0}$ &  $\sigma^2_\mathcal{T}$    & \redcross \\
        \textbf{CCDF}        & $\alpha_{\tau^\prime} g(\y^{t+1})$         & $\sigma_{\tau^\prime}^2$ & \redcross \\
        \textbf{SeqDiff}     & $\alpha_{\tau^\prime}\x_0^t$               &$\sigma_{\tau^\prime}^2$ & \orangetilde \\
        \textbf{SeqDiff+}     & $\alpha_{\tau^\prime}\textcolor{purple}{f_\theta(\x_0^{t-K:t})}$   &$\sigma_{\tau^\prime}^2$ & \greencheck \\
        \bottomrule
    \end{tabular}
    \caption{Comparison of the different initialization methods for accelerating reverse diffusion trajectories.}
    \label{tab:methods_comparison}
    \vspace{-0.4cm}
\end{table}

\subsection{SeqDiff}
One straightforward method of initialization given past past observations is to directly use the previous diffusion posterior estimate $\x_0^t$ as an estimate for the mean of $\x^{t+1}$. This would lead to the following diffusion initialization for $t+1$:
\begin{equation}
    \x_{\tau^\prime}^{t+1} \sim p(\x_{\tau^\prime}^{t+1} \mid \x_0^{t+1}) \approx \mathcal{N}(\alpha_{\tau^\prime}\x_0^t, \sigma_{\tau^\prime}^2 \I).
\label{eq:seqdiff}
\end{equation}
This assumes a simple linear sequential model, which we show is reasonable in case of high frame rate scenarios where the temporal correlation across subsequent frames is strong.

\subsection{SeqDiff+}
In cases of severe motion or lower frame rates we leverage a ViViT network $f_\phi(\cdot)$ to model the system dynamics and predict the mean of the next frame for improved initialization. This allows us to improve on \eqref{eq:seqdiff}, as follows:
\begin{equation}
    \x_{\tau^\prime}^{t+1} \sim p(\x_{\tau^\prime}^{t+1} \mid \x_0^{t+1}) \approx \mathcal{N}(\alpha_{\tau^\prime}\tilde{\x}_0^{t+1}, \sigma_{\tau^\prime}^2 \I),
\end{equation}
where $\tilde{\x}_0^{t+1}$ is predicted by the transformer model $f_\phi$, parameterized with $\phi$, which takes as input a sequence of past posterior estimates and outputs a prediction of the next frame as follows:
\begin{equation}
    \tilde{\mathbf{x}}_0^{t+1} = f_\phi(\x_0^{t}, \x_0^{t-1}, ...,  \x_0^{t-K}).
\end{equation}
A full comparison of all diffusion initialization methods is listed in Table~\ref{tab:methods_comparison}, and illustrated in Fig.~\ref{fig:schematic}
\begin{figure*}
    \centering
    \includegraphics[width=1\linewidth]{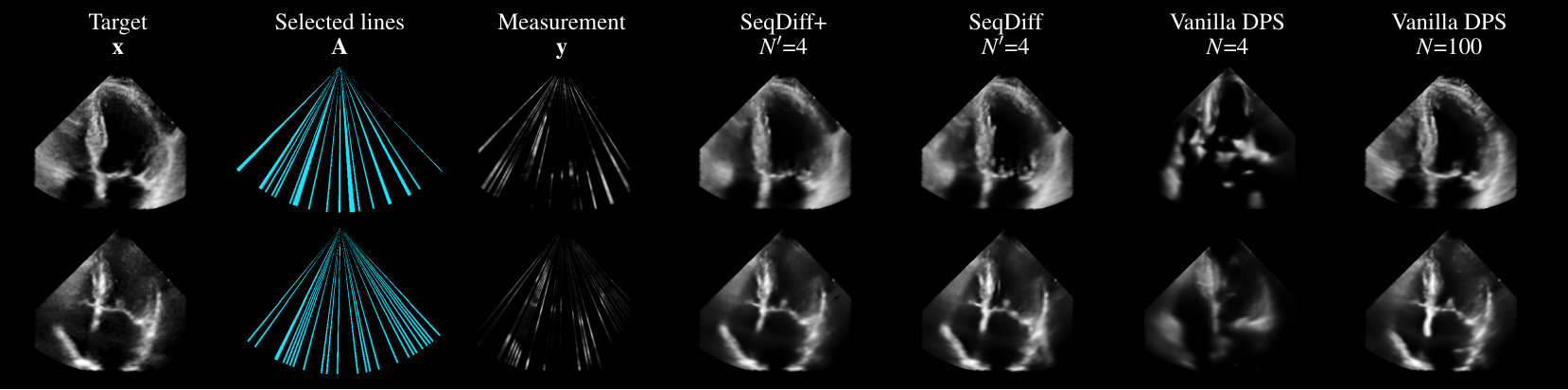}
    \caption{Qualitative comparison of Vanilla DPS (for $N=4$ and $N=100$ steps), and the two proposed initialization methods SeqDiff and SeqDiff+ for only $N^\prime=4$ diffusion steps. Target images $\x^t$ are 80\% masked by $\A^t$ to produce observation $\y^t$. Initialization with SeqDiff(+) is able to improve on full diffusion trajectories with $25\times$ speedup.}
    \label{fig:comparison}
\vspace{-0.5cm}
\end{figure*}

\begin{figure*}[h]
    \newcommand{\labelraise}{0.6cm}
    \centering
    \begin{subfigure}[b]{0.32\textwidth}
        \centering
        \includegraphics[trim=0cm 0.2cm 0cm 0.2cm, clip, width=1\textwidth]{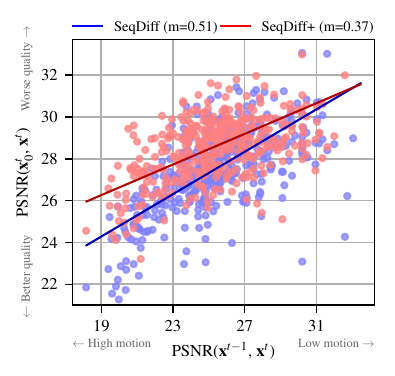}
        \captionsetup{justification=raggedright,singlelinecheck=false, font=bf, skip=-\labelraise}
        \caption{}
        \label{fig:motion-error}
    \end{subfigure}
    \hfill
    \begin{subfigure}[b]{0.32\textwidth}
        \centering
        \includegraphics[trim=0cm 0.2cm 0cm 0.2cm, clip, width=1.04\textwidth]{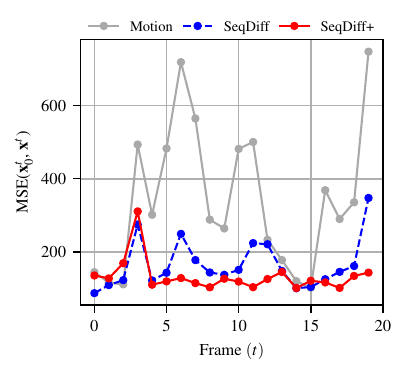}
        \captionsetup{justification=raggedright,singlelinecheck=false, font=bf, skip=-\labelraise}
         \caption{}
        \label{fig:motion-sequence}
    \end{subfigure}
    \hfill
    \begin{subfigure}[b]{0.32\textwidth}
        \centering
        \includegraphics[trim=0cm 0.1cm 0cm 0.3cm, clip, width=1\textwidth]{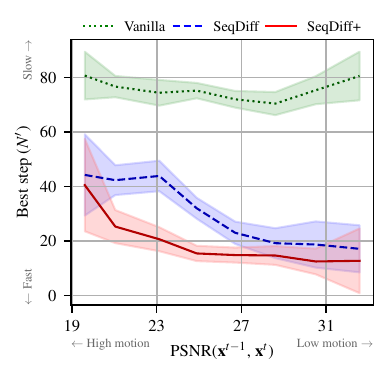}
        \captionsetup{justification=raggedright,singlelinecheck=false, font=bf, skip=-\labelraise}
        \caption{}
        \label{fig:best-step}
    \end{subfigure}
    \captionsetup{skip=10pt}
    \caption{Comparison of SeqDiff(+) performance in PSNR against various motion conditions. \textbf{(a)} For every sample in the test set. The advantage of using a transition model (SeqDiff+) is most advantageous with high motion (see linear fit $m$). \textbf{(b)} For a single sequence of frames. SeqDiff+ is less correlated with the motion, whereas the error of SeqDiff increases with more movement, emphasizing the importance of the transition model. \textbf{(c)} Best performing $N^\prime$ for each initialization method against motion. SeqDiff+~outperforms the other methods for all motion levels. For lower motion levels, SeqDiff is a valid option.}
    \label{fig:combined}
    \vspace{-0.5cm}
\end{figure*}

\section{Results}
\label{sec:results}
\begin{figure}
    \centering
    \includegraphics[trim=0cm 0.2cm 0cm 0.2cm, clip, width=0.85\linewidth]{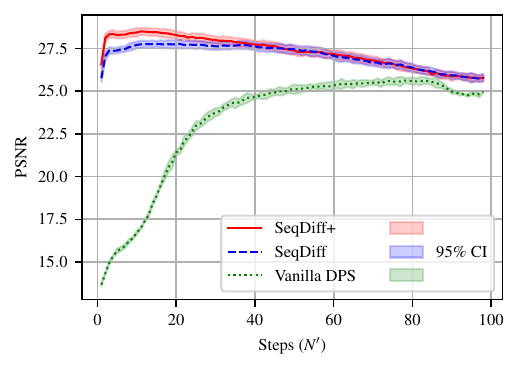}
    \caption{PSNR against number of diffusion steps on sequences of frames from the test split of EchoNet-Dynamic dataset. Confidence Interval (CI) is taken over 3 splits with different masks and seeds. SeqDiff+ shows a notable improvement, particularly with fewer diffusion steps $N^\prime$.}
    \label{fig:psnr}
\vspace{-0.4cm}
\end{figure}

To evaluate our methods, we test conditional diffusion trajectories (vanilla DPS) with and without SeqDiff and SeqDiff+ initialization strategies on the EchoNet-Dynamic dataset~\cite{ouyang2020video} with approximately 7000 sequences of around 80 to 300 frames each of which we reserve 100 sequences for evaluation. We map all images to a polar grid to retrieve the original scanning lines and resize to $128\times128$. After inference the images are scan converted back to cartesian grid for display and metrics calculation. Subsampling is a compressed sensing technique frequently used in medical imaging to reduce data rates~\cite{huijben2020learning, bakker2020experimental, stevens2022accelerated, nolan2024active}. As a reconstruction task for the diffusion model we consider a scan-line undersampling task which can be used in ultrasound imaging to reduce acquisition time and is essentially subsampling of the image columns. For SeqDiff+ we use a ViViT architecture with context length $K=4$, two transformer layers with 8 heads for encoder and decoder each and a tublet size of $(2, 16, 16)$. Unless specified otherwise, results are generated with only $N^\prime=4$ diffusion steps. In Fig.~\ref{fig:comparison} a visual comparison of the initialization methods is shown, with the proposed methods clearly outperforming both vanilla DPS given the same number of diffusion steps, as well as full diffusion trajectory with $25\times$ fewer steps. This is reflected in the metrics too, as seen in Fig.~\ref{fig:psnr}, where SeqDiff+ initialization outperforms its counterpart without transition model, especially for low $N^\prime$. For high $N^\prime$ the performance tapers off as useful past information is \emph{forgotten} due to the noise being added. The importance of an accurate transition model is highlighted in Fig.~\ref{fig:motion-error}, Fig.~\ref{fig:motion-sequence} and Fig.~\ref{fig:best-step} which compare the performance against motion. We observe that in cases with higher motion it pays off to use the ViViT to account for the dynamics. Furthermore, based on the amount of motion, SeqDiff(+) offers a way to determine the optimal initialization point $\tau^\prime$ as seen in Fig.~\ref{fig:best-step}.

\section{Conclusions}
\label{sec:conclusions}
In this paper, we introduce a novel sequential posterior sampling approach, coined SeqDiff(+), to accelerate diffusion models in the context of sequence data. Our method capitalizes on the temporal structure between subsequent frames which enables autoregressive sampling based on previous posterior estimates. Additionally, we adapt a Video Vision Transformer (ViViT) to model the transition dynamics between frames for improved initialization of the diffusion process. Our approach effectively reduces the number of diffusion iterations with respect to full conditional diffusion trajectories up to $25\times$, unlocking the use of diffusion models for real-time imaging applications such as ultrasound imaging. We evaluate our approach on scan-line undersampling in cardiac ultrasound frames and show that, especially in cases with severe motion, the addition of a transition model further improves performance.

\vfill\pagebreak

\clearpage

\bibliographystyle{IEEEtran}
\bibliography{refs}

\balance

\end{document}